\documentclass{article}
\usepackage{ijcai17}

\usepackage{times}

\usepackage{amssymb}
\usepackage{amsmath}
\usepackage{graphicx,subfigure}
\usepackage{textcomp}
\usepackage{multirow,url}

\title{Autoencoder Regularized Network For Driving Style Representation Learning%\thanks{These match the formatting instructions of IJCAI-07. The support of IJCAI, Inc. is acknowledged.}
}
\author{Weishan Dong, Ting Yuan, Kai Yang, Changsheng Li, Shilei Zhang\\
IBM Research -- China\\
\normalsize \{weishan.dong, iduduyuan, satisfiedkai\}@gmail.com, 
changsheng\_li\_507@hotmail.com, slzhang@cn.ibm.com}

\begin{document}

\maketitle

\begin{abstract}
In this paper, we study learning generalized driving style representations from automobile GPS trip data. % collected by automobile GPS sensors.
We propose a novel Autoencoder Regularized deep neural Network (\emph{ARNet}) and a trip encoding framework \emph{trip2vec} to learn drivers' driving styles directly from GPS records, by combining supervised and unsupervised feature learning in a unified architecture. %, resulting better generalized performance on new drivers that have not appeared in the training set.
Experiments on a challenging driver number estimation problem and the driver identification problem show that ARNet can learn a good generalized driving style representation: It significantly outperforms existing methods and alternative architectures by reaching the least estimation error on average (0.68, less than one driver) and the highest identification accuracy (by at least 3\% improvement) compared with traditional supervised learning methods.
\end{abstract}

\section{Introduction}
\label{sec:Intro}

Studying human drivers' driving behaviors from automobile sensor data is an interesting research topic.
Similar to biometrics such as gait, voice, and typing rhythm, %due to individual differences,
each driver also has a signature pattern of driving, %, resulting more or less different driving behaviors even if driving a same car on a same route
which is also called driving style \cite{lin2014overview}.
There are many aspects that can be used to measure driving styles. In this paper, we focus on studying vehicle movement measures including speed change, turning, and their temporal combinations derived from GPS (Global Positioning System) sensor data that are collected in a short and regular time interval (e.g., 1 second). These measures can reflect drivers' fine-grained behavioral habits of steering and speed control. % under certain road conditions.

Learning driving style representations from automobile sensor data %for driver identification
has been intensively studied \cite{van2013driver,lin2014overview,kuderer2015learning,dong2016characterizing}. Compared with other automobile sensors, such as OBD (On-Board Diagnostic) system, CAN (Controller Area Network) buses and cameras, GPS sensor data are often easier to collect, % due to its less intrusiveness and concerns from privacy and safety,
making them popular in large-scale research.
Auto insurance companies have become highly interested in utilizing driving style information extracted from GPS data to solve their business problems \cite{laurie2011telematics}. %,kaggle2015telematics}. 
A good driving style representation can help answer questions such as if the driver identified on the policy is driving the proper car, how many drivers share a car, and if an additional driver is driving a car, etc.
Insights to these questions can be critical for risk evaluation and can be applied to policy premiums in insurance programs such as pay-as-you-drive. % \cite{desyllas2013profiting}.
Besides, a good driving style representation also helps to better modeling and understanding human drivers' behaviors, which are beneficial to improving the designs of driving assistance systems, driver-car interactions, and autonomous driving \cite{lin2014overview,kuderer2015learning}.

%Due to its business value, lots of research work have been done.
Existing approaches typically follow the supervised learning paradigm, where the inputs are vehicle movement features derived from GPS data and the labels are drivers' identities.
The learning process is usually guided by minimizing a classification loss.
The driving style representation (features) learnt in this way can work well in describing unseen trips of seen drivers. Nonetheless, %due to the i.i.d. assumption,
the learnt representation is not guaranteed to be a good generalized representation on \emph{unseen drivers}.
When the number of drivers in the training set is small, the learnt model can hardly work well since unseen drivers' driving behaviors can be extremely diverse in practice.
On the other hand, collecting data from a large number of drivers and ensuring a sufficiently large trip training set for each driver can be challenging. Besides, when the number of drivers becomes large (say, thousands), the classifier can become difficult to train.

\begin{figure}[!t]
\centering
\includegraphics[width=0.85\linewidth]{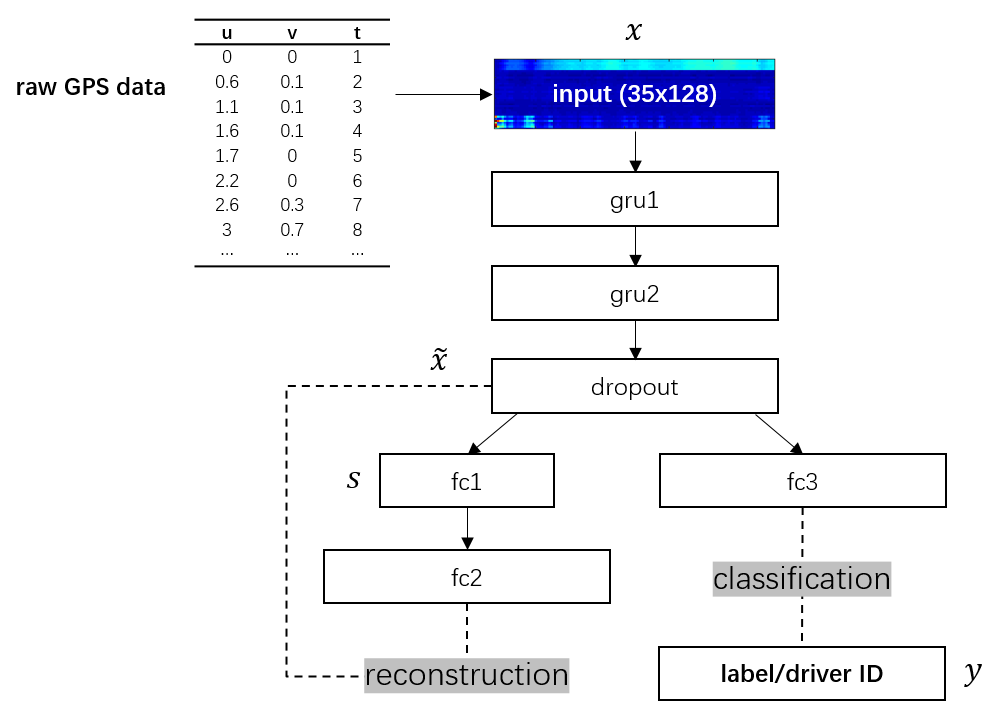}
\vspace{-0.1in}
\caption{ARNet architecture for driving style representation learning.
%Number of hidden units is shown in brackets.
Dash lines indicate supervisory information.}
\label{fig:architecture}
\vspace{-0.15in}
\end{figure}

To solve the problem, in this paper, we propose a novel deep neural network, {A}utoencoder {R}egularized {Net}work (\emph{ARNet}) for generalized driving style representation learning.
%Compared with traditional methods, based on the same training data from known drivers, it learns driving style features that work significantly better on describing unseen drivers.
Figure~\ref{fig:architecture} illustrates the overall architecture.
Different from existing deep neural networks, ARNet directly learns from GPS data and combines supervised and unsupervised feature learning in one architecture.
The motivation is to use a specially designed autoencoder structure to regularize the discriminative feature learning in a classification network.
In ARNet, %an autoencoder for unsupervised feature reconstruction is applied
supervised learning is combined with unsupervised feature reconstruction using a Recurrent Neural Network (RNN) \cite{elman1990finding,chung2014empirical} output as a shared hidden layer.
An $l_1$ regularized bottleneck layer (fc1 in Figure~\ref{fig:architecture}) of the autoencoder structure %for the reconstruction task
serves as the final driving style feature representation extraction layer.
The feature learning is guided simultaneously by a classification loss defined on trip labels and a reconstruction loss defined on how well the driving style feature layer reconstructs the hidden RNN output.
Notably, the autoencoder in ARNet aims to reconstruct the hidden-layer RNN feature that keep changing in training, but not to reconstruct the fixed network inputs as in typical autoencoder networks.
Such a design can be viewed as a form of regularization to the hidden-layer RNN feature for classification: The feature should be discriminative, meanwhile, they should be reconstructible.
The bottleneck layer of the autoencoder (fc1) thus can learn the basis of driving styles, which is expected to generalize better on unseen drivers.
Reversely, the design can also be regarded as introducing supervisory information to unsupervised feature learning:
Labels of limited training samples bring in prior knowledge to the unsupervised autoencoder, making the learnt basis feature more meaningful and discriminative.
%%The network reads transformed GPS data as inputs.
%Unlike most existing multi-task learning neural networks that combine multiple tasks of a same type, i.e., either all supervised tasks or all unsupervised tasks, in our architecture, a supervised classification task is combined with an unsupervised reconstruction task using a recurrent neural network (RNN) \cite{elman1990finding,chung2014empirical} output as a shared layer.
%An $l_1$ regularized bottleneck layer (fc1 in Figure~\ref{fig:architecture}) of an autoencoder sub-architecture %for the reconstruction task
%serves as the driving style feature extraction layer.
%The feature learning is guided simultaneously by a classification loss defined on trip labels and a reconstruction loss defined on how well the driving style feature layer reconstructs the RNN output.
%%It can be proved that such a representation learning is the best convex approximation to additionally applying K-means to the RNN extracted features, resulting a sparse representation generalized better on unseen drivers.
Furthermore, we also propose a trip encoding framework, namely \emph{trip2vec}, which encodes a varied-length GPS trip into a fixed-length vector describing the trip-level driving style using the proposed ARNet as the base encoder.
%Such a representation is robust to factors such as local road shapes that may influence short-term driving behaviors.

%To demonstrate the advantage of ARNet,
We study two problems as benchmarks on a large real dataset.
The first one is a new and challenging real-world problem raised from the auto insurance industry, called driver number estimation.
The objective is to identify the true number of drivers from a set of anonymous trips.
Importantly, these drivers are new and unseen to the training phase.
%By applying a clustering algorithm to the encoded driving style representations, we can estimate how many drivers generated the trips.
Large-scale experimental studies show that ARNet significantly outperforms alternative methods.
On a wide range of tests (from 1 to 10 drivers), the average absolute error between the estimation and the ground truth is just 0.68 (less than one driver). In contrast, other candidate methods lead to errors much larger than 1.
The second problem is the classical driver identification problem, measured by the classification accuracy on unseen trips of seen drivers. Experiments on a 50-class problem show that, ARNet reaches the highest classification accuracy by at least 3\% improvement compared with several existing classification-based methods.% on a 50-class problem.
%These validate the effectiveness of multi-task learning: It benefits from both tasks as well as improves the performance on each task.

%The contribution of this paper are as follows. First, we propose a novel multi-task neural network architecture, which combines supervised and unsupervised learning.

\section{Autoencoder Regularized Network (ARNet) And Trip2vec Encoding}
We first introduce ARNet that reads GPS data as inputs and learns a compact driving style feature representation. % by combining supervised classification and unsupervised reconstruction.
Then, we introduce trip2vec, a trip encoding framework, which extends the learnt driving style representation to trip-level.

\subsection{GPS Data Transformation}
A trip (i.e., GPS trajectory) can be defined by a varied-length sequence of tuples ($u, v, t$), where ($u, v$) denotes a geo-location and $t$ denotes time.
We follow the data transformation method proposed by \cite{dong2016characterizing} to construct neural network inputs from raw GPS data, which has been proved an effective way of extending deep learning to working on GPS data.
A trip is first windowed into segments of a fixed length \(L_s\) with a shift $L_s/2$, each of which encodes five instantaneous car movement features, namely basic features, derived from neighboring GPS data: speed norm, difference of speed norm, acceleration norm, difference of acceleration norm, and angular speed.
Then, each segment is further applied a sliding window of length \(L_f\) ($L_f < L_s$) with a shift $L_f/2$. Each window produces a frame encoding seven statistics of the basic features: mean, minimum, maximum, 25\%, 50\% and 75\% quartiles, and standard deviation.
As a result, a set of statistical feature matrices of $5 \times 7=35$ rows and $2\times L_s/L_f$ columns can be obtained from a given trip.
A feature matrix describing one trip segment defines an input sample to neural networks.
For instance, given GPS data sampled once per second as in the experiments, using $L_s=256s,L_f=4s$, we define feature matrices of size $35\times 128$ as network inputs.
The label of a sample inherits from the trip to which the corresponding segment belongs.

\subsection{Autoencoder Regularized Network (ARNet)}
The proposed ARNet architecture is depicted in Figure~\ref{fig:architecture}, which consists of three parts: a stacked RNN, an autoencoder for reconstruction, and a softmax for classification.

\subsubsection{Stacked RNN}
Let $x$ denote the $35\times 128$ input, i.e., a trip segment.
A stacked RNN (gru1+gru2+dropout in Figure~\ref{fig:architecture}) reads $x$ to extract higher-level features.
As driving style is typically the temporal combination of driving actions, % (see Section~\ref{sec:Intro}),
we regard $x$ as a sequence of length 128 and each element of which as a 35-d vector.
Here we employs a 2-layer stacked GRU (Gated Recurrent Unit) \cite{cho2014properties} architecture to exploit the sequential dependencies.
GRU network has been proved an effective RNN design \cite{chung2014empirical}.
Our empirical studies show that GRU works slightly better than several popular RNN architectures on our problem, including LSTM \cite{hochreiter1997LSTM} and bi-directional RNN \cite{schuster1997bidirectional}, hence we adopt GRU in the design.
%The first GRU layer (gru1) contains 256 hidden units. It reads $35\times 128$ input $x$ with unrolling itself 128 steps along the time axis, and outputs a sequence of a same length 128, but each element of which is a 256-d vector.
%The second GRU layer (gru2), also containing 256 hidden units, is appended to gru1. It also unrolls 128 steps, but outputs a 256-d vector instead of a sequence.
%A dropout layer with probability 0.5 is applied to gru2 to reduce overfitting \cite{hinton2012improving}.
%It plays as a shared feature layer bridging the two tasks.
%Let $\tilde{x}$ denote its 256-d output given an input $x$.
The first GRU layer (gru1) reads $35\times 128$ input $x$ with unrolling itself 128 steps along the time axis, and outputs a sequence of a same length 128, each element of which is a vector.
The size of the vector (i.e., dimension) equals to the number of hidden units in gru1.
The second GRU layer (gru2) is appended to gru1. It also unrolls 128 steps, but outputs a vector instead of a sequence.
The size of the vector equals to the number of hidden units in gru2.
A dropout layer is applied to gru2 to reduce overfitting \cite{hinton2012improving}.
This dropout layer plays as the shared hidden feature layer bridging the supervised and unsupervised learning.
Let $\tilde{x}$ denote its output given an input $x$.

\subsubsection{Autoencoder}
A 3-layer autoencoder (dropout+fc1+fc2 in Figure~\ref{fig:architecture}) is employed for feature reconstruction.
Notably, we use the autoencoder to reconstruct $\tilde{x}$ instead of $x$, which is critical for learning better generalized driving style representation.
A fully-connected bottleneck layer (fc1) %with 50 hidden units
is used to learn a compressed representation $s$ of $\tilde{x}$.
ReLU nonlinearity $f(z)=\max(0,z)$ is used in fc1 to ensure $s$ non-negative, which will be used in the trip2vec encoding (see Eq (\ref{eq:trip_encoding})).
$l_1$ sparsity regularization is applied on $s$ (see Eq (\ref{eq:reconstruct_loss})). %Section~\ref{sec:Objective}.
A fully-connected layer (fc2) %with 256 units
is the output layer of the autoencoder, where $f(z)=\tanh(z)$ activations are used to approximate $\tilde{x}$ for reconstruction.
%Let $\mathcal{D}$ denote the overall transformation applied on $s$ in fc2. Then fc2 output is $\mathcal{D}\cdot s$, which approximates $\tilde{x}$ for reconstruction.
%The fc2 layer output is used to approximate $\tilde{x}$ for reconstruction.

\subsubsection{Softmax Regression}
A fully-connected layer (fc3 in Figure~\ref{fig:architecture}) appending to the dropout layer is defined for classification. A softmax regression is applied to produce a distribution over class labels.
The number of classes (denoted by $c$) equals to the number of drivers in the training set.

\subsubsection{Objective Function and Approximation}
\label{sec:Objective}
Given a training set $\{x_i, y_i\}$ where $i\in\{1,\ldots,n\}, y_i\in \{1,\ldots, c\}$, the overall objective function is defined as a combination of reconstruction and classification objectives.
The reconstruction loss is defined as:
\vspace{-0.05in}
\begin{equation}\label{eq:reconstruct_loss}
%\small
%\min
\mathcal{J}_r =
\sum_{i}^n||\mathcal{D} s_i - \tilde{x}_i||_2^2 + \lambda ||s_i||_1
\vspace{-0.05in}
\end{equation}
where $\tilde{x}_i$ is the output of the RNN dropout layer, $\mathcal{D}\in \mathcal{R}^{(n\times k)}$ is a ``dictionary" of $k$ vectors,
$s_i\in \mathcal{R}^k$ is a ``code vector" associated with the $\tilde{x}_i$.
%Note that $k=50$ in Figure~\ref{fig:architecture}.
The first term of $\mathcal{J}_r$ is the reconstruction error, which intends to find a dictionary $\mathcal{D}$ and a new representation $s_i$ to reconstruct $\tilde{x}_i$, the learnt feature from RNN. $l_1$ regularization is used to encourage $s_i$ to be sparse. Eq~(\ref{eq:reconstruct_loss}) is a sparse coding objective \cite{sparsecoding}.
The classification loss is defined as the standard cross-entropy:
\vspace{-0.05in}
\begin{equation}\label{eq:classify_loss}
%\small
%\min
\mathcal{J}_c =
-\frac{1}{n}\sum_{i}^n\sum_j^c 1\{y_i=j\} \log\frac{e^{\theta_j^T \tilde{x}_i}}{\sum_l^c e^{\theta_l^T \tilde{x}_i}}
\vspace{-0.05in}
\end{equation}
%\vspace{-0.1in}
where $1\{\cdot\}$ is the indicator function and $\theta = \{\theta_1,\ldots,\theta_c\}$ are the softmax regression parameters.
Overall, the combined objective function is defined as:
%\vspace{-0.05in}
\begin{equation}\label{eq:multitask_obj}
%\small
\min \mathcal{J}_r + \mathcal{J}_c
%\vspace{-0.05in}
\end{equation}
%\vspace{-0.1in}

The motivation of Eq~(\ref{eq:reconstruct_loss}) is as follows.
%learning $s$ to approximate $\tilde{x}$ instead of $x$ in
If excluding the autoencoder layers (fc1+fc2), the network becomes a stacked RNN and $\tilde{x}$ can be used as a feature representation of $x$.
In this way, the learning of $\tilde{x}$ is guided only by supervisory information of trip labels (i.e., driver IDs). The dropout can help reduce overfitting, but considering the number of \emph{unseen} drivers can be extremely large, given limited training data, the learning is still prone to overfit to the \emph{seen} drivers in the training set.
Therefore, $\tilde{x}$ can hardly be a good representation for unseen drivers.
As a straightforward extension, we want to learn a representation which has the clustering characteristic and be more compact to have better generalization performance. Minimizing $\mathcal{J}_r$ can help to achieve this goal.
Let us look at the objective of a classical clustering algorithm, K-means \cite{kmeans2012learningfeature}:
%\vspace{-0.05in}
\begin{equation}\label{eq:kmeans_obj}
%\small
\min_{\mu} \sum_k  (|| x_i - \mu_k ||^2)
%\vspace{-0.05in}
\end{equation}
%\vspace{-0.1in}
which intends to find cluster centroids $u_k$ that minimize the distance between data points and the nearest centroid.
Equally to Eq~(\ref{eq:kmeans_obj}), K-means can be viewed as a way of reconstructing $x_i$ \cite{kmeans2012learningfeature}:
%\vspace{-0.05in}
\begin{align}\label{eq:kmeans_obj2}
%\small
\min_{\mathcal{D},s} \sum_i(|| \mathcal{D}s_i - x_i ||^2), \quad%\nonumber\\
s.t. \hspace{0.05in}  || s_i||_0 \leq 1, \forall i
%\vspace{-0.05in}
\end{align}
%\vspace{-0.1in}
Compare Eq~(\ref{eq:reconstruct_loss}) and Eq~(\ref{eq:kmeans_obj2}), they optimize the same type of reconstruction objective. The only difference is that Eq~(\ref{eq:reconstruct_loss}) allows more than one non-zero entry in each $s_i$, enabling a much more accurate representation of each $x_i$ while still requiring each $s_i$ to be simple.
Therefore, minimizing $\mathcal{J}_r$ can make the learnt representation have clustering characteristics and be more compact. % to have better generalization performance.

To achieve the goal of minimizing Eq~(\ref{eq:reconstruct_loss}), in ARNet, we use the layers dropout+fc1+fc2 to approximate so as to make the unified architecture easier to train. Known as the sparse coding objective, Eq~(\ref{eq:reconstruct_loss}) intends to learn a sparse reconstruction and an efficient coding for $\tilde{x}_i$, which shares the spirit of sparse autoencoders \cite{ng2011sparse}.
The dropout+fc1+fc2 is a typical autoencoder structure, and we use $l_1$ regularization on the output of fc1 so that the coding for $\tilde{x}_i$ is sparse.

We can read ARNet from two perspectives.
On one hand, we can view $\mathcal{J}_r$ (Eq~(\ref{eq:reconstruct_loss})) as regularization terms if regarding $\mathcal{J}_c$ (Eq~(\ref{eq:classify_loss})) as the main loss.
The autoencoder structure plays as a regularizer to the classification feature $\tilde{x}$, and should improve classification performance.
On the other hand, we can view $\mathcal{J}_c$ as a term introducing prior knowledge to the unsupervised feature learning guided by $\mathcal{J}_r$.
The finally learnt feature $s$ thus should work better than features learnt from purely unsupervised learning.
Experimental studies in the next section will verify these from both sides.

\subsection{Trip2vec: A Trip Encoding Framework}
\label{sec:TripEncoding}

\begin{figure}[tb]
\centering
\includegraphics[width=0.97\linewidth]{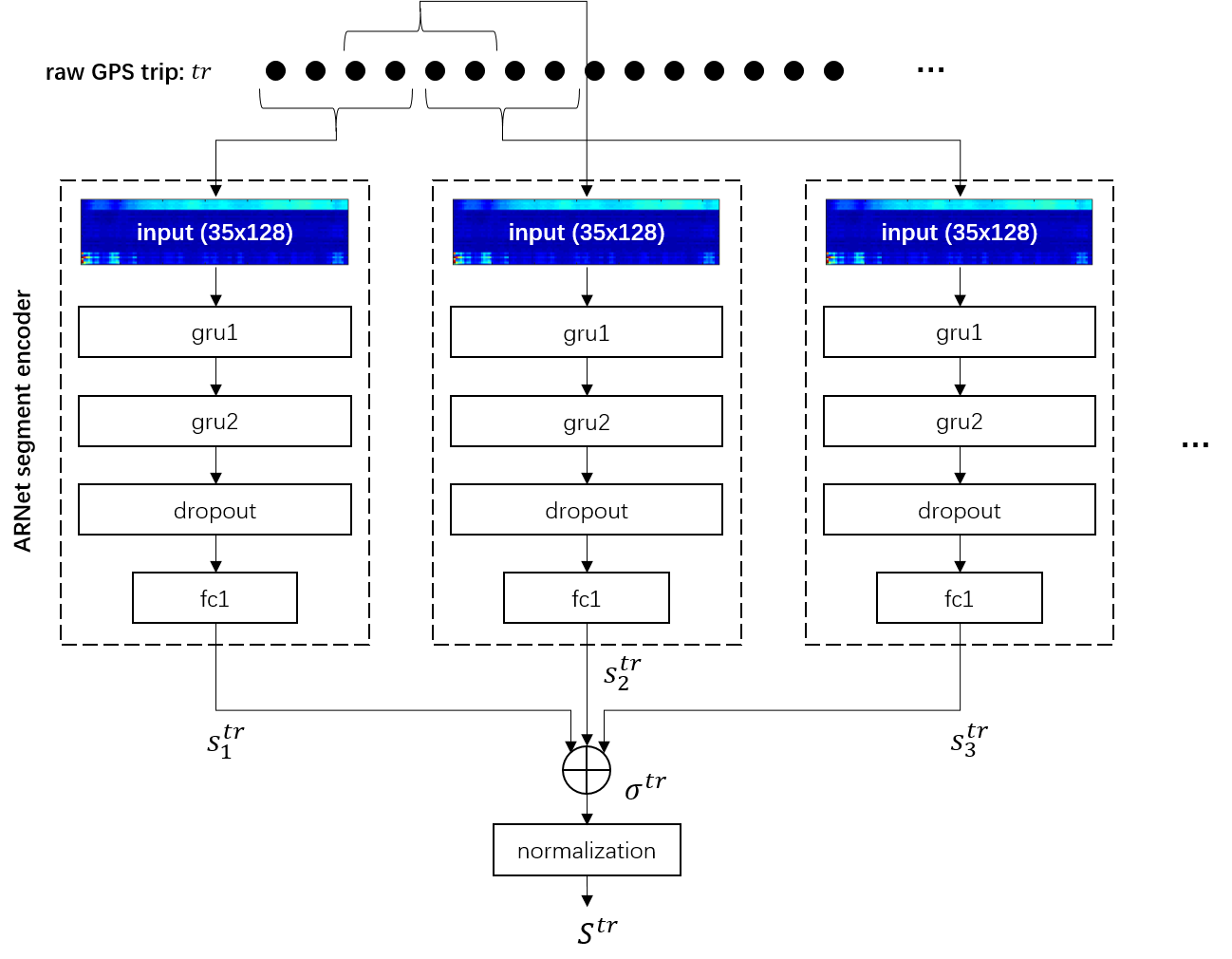}
\vspace{-0.15in}
\caption{Trip2vec framework}
\label{fig:trip2vec}
\vspace{-0.1in}
\end{figure}

Once we have trained the ARNet, we can use the layers gru1+gru2+dropout+fc1 as a trip segment encoder. %, which outputs a 50-d driving style feature vector given a segment input.
But still, it extracts driving style information only from trip data on segment-level, which can be impacted by local factors such as road shapes and traffic conditions, etc.
Therefore, we need a more robust representation that describes trip-level driving styles.
In order to do so, we propose a trip encoding framework, called \emph{trip2vec}, which adopts the Bag-of-Words (BoW) feature construction strategy \cite{fei2005bayesian} based on the trained trip segment encoder.
Roughly speaking, we can treat a varied-length trip as an ``article'' and each segment as a ``paragraph''.
The overall ``topic'' of an article can be derived from aggregating the paragraph-level information.
Similarly, based on the segment-level driving styles, we propose to define the trip-level driving style representation by the normalized sum of all the segment-level feature vectors.
Figure~\ref{fig:trip2vec} illustrates the trip2vec framework.
Suppose a trip $tr$ is divided into $q$ segments%$\{x^{tr}_i\}, i\in\{1,\ldots,q\}$
, and the encoded segment features are %$\{s^{tr}_1, s^{tr}_2, \ldots, s^{tr}_q\}$.
$\{s^{tr}_i\}, i\in\{1,\ldots,q\}$.
The trip-level driving style feature representation is defined as:
%\vspace{-0.05in}
\begin{equation}\label{eq:trip_encoding}
%\small
S^{tr} = \frac{\sigma^{tr}}{\max_j \{\sigma_{j}^{tr}\}}
%\vspace{-0.05in}
\end{equation}
%\vspace{-0.1in}
%where $\sigma=\sum_i^q s^{tr}_i$.
where $\sigma^{tr}=\sum_i^q s^{tr}_i$ is the vector sum, and $\sigma_j^{tr}$ denotes its $j$-th dimension ($\sigma_{j}^{tr}>=0, \forall tr, j$).

\section{Experiments}
\label{sec:Experiments}
We use a large real yet private dataset %from the Kaggle 2015 competition on driver telematics analysis \cite{kaggle2015telematics} 
in experiments.
The dataset is collected by an insurance company, containing
%contains
over 500,000 %anonymized 
GPS trips from over 2,500 drivers.
Each driver has 200 trips that record the car's location every second.
%\footnote{The dataset contains a small number of false trips that actually do not belong to the labeled driver, but unfortunately the ground truths are not available. In our experiments, we assume the labels are accurate and regard the false trips as noise.}.
%We further processed the data to exclude trips that either are too short (less than $L_s=256s$) or contain huge GPS errors (outliers with instantaneous speed exceeding 70m/s, equally 252km/h).

\subsection{Driver Number Estimation Problem}
We first study the driver number estimation problem. % to demonstrate the advantages of multi-task driving style feature learning.
The aim is to estimate the number of drivers from a set of anonymous trips.
The driving style representation learning is based on labeled trips of a set of known drivers, but the testing trips are from {unseen} new drivers. %, meaning the labels are not even included in the training set.
This is to mimic the situations in real-world, where the auto insurance companies are interested to know how many drivers share a car given this car's recorded trips. %, which is a risk factor.
However, the driver ID is unknown for the trips.
More importantly, the potential drivers are most likely new, meaning their data are not available in model training, which makes the problem a challenging one.
A precise estimation can help improve the risk modeling and pricing policies and to generate direct business values.

\subsubsection{Experimental Settings}
For comparisons, we include two alternative architectures: a reconstruction-only network (RONet) and a classification-only network (CONet), %denoted by RONet and CONet, respectively,
which are defined by removing one of the two losses from ARNet. % in Figure~\ref{fig:architecture}.
For all the nets, we set 256 hidden units in gru1 and gru2, thus dropout output $\tilde{x}$ and fc2 output are both 256-d.
We use dropout probability 0.5.
We set 50 hidden units in fc1, thus the final driving style representation learnt by ARNet is a 50-d vector.
We use $\lambda$=1e-5, ADADELTA optimizer \cite{zeiler2012adadelta} with learning rate 1.0, $\rho$=0.95 and $\epsilon$=1e-8, and batch size 2560 in training these networks.
The training is based on the trip data from the first 50 drivers in the dataset.
For each driver, we use 80\% trips as training data, and the rest 20\% as classification validation data.
Training ARNet and CONet stops until the %segment-level 
validation accuracy is maximized (at epochs 33 and 116, respectively).
Training RONet stops when the reconstruction loss $\mathcal{J}_r$ converges around 0.001
(at epoch 100).
We use fc1 as the driving style feature layer for ARNet and RONet, and use the dropout layer for CONet.
Trip features are computed using the proposed trip2vec framework %from the obtained segment-level features 
for all the nets.

We also include a 57-d handcrafted trip feature representation proposed by \cite{dong2016characterizing} as another baseline, which demonstrated good classification performance working with GBDT (Gradient Boosting Decision Tree) \cite{friedman2001greedy}.
We denote it by TripGBDT feature.
The 57 features include global and local driving behavior statistics.
The global ones are trip-level statistics of speed, acceleration, and angular speed, total trip time duration, total trip length, trip average speed, and size of the minimal bounding rectangle describing trip geometry.
The local ones are statistics of movement features calculated on different time scales with correlation to binned local road shapes.
%Readers can refer to \cite{dong2016characterizing} for details of the feature definitions.
We will compare it with those trip features learnt by the deep architectures.

Based on the trip features, we employ Affinity Propagation (AP) \cite{frey07affinitypropagation} to cluster the trips and estimate the number of drivers.
Assuming different drivers should have different driving styles, each obtained cluster refers to the trips belong to one driver, and thus the number of clusters reflects the number of drivers.
AP has the advantage of automatically determining the number of clusters.
We employ the scikit-learn implementation of AP \cite{scikit-learn} using the Euclidean affinity, % (the negative squared Euclidean distance), 
where a preference parameter is needed.
We tuned this parameter for each candidate, and chose preference values -5, -8.5, -12, and -3.5 for ARNet, RONet, CONet, and TripGBDT features, respectively.
This is to reduce the effect of clustering algorithm and to fairly compare different feature representations.
We use the default damping factor 0.5 in AP, since empirical studies showed that the results are insensitive to this parameter.

\subsubsection{Design of Testing}
%The testing is done as follows.
We sample from the unseen drivers (ID greater than 50) who are absent in training to construct testing sets.
We build 10 test groups. Each corresponds to a fixed number of drivers, ranging from 1 to 10.
%Each driver has about 160 trips on average. %, thus the amount of trips of a test group increases approximately linearly with the number of drivers.
For each test group, we randomly sample 25 times from the unseen drivers.
%Once a driver is chosen, all his/her trips are pu
As a result, each test group contains 25 trip sets, and each trip set refers to a random combination of drivers.
%We record the performance of candidate algorithms on all the test groups to evaluate the robustness of the driving style features learnt.
We compute two metrics: (1) the absolute error between the true number of drivers and the estimates, and (2) the AMI (Adjusted Mutual Information) score \cite{vinh2010information} measuring the clustering quality, which returns a value of 1 when the cluster partitions are perfectly matched with the true labels, while random partitions have an expected AMI around 0.
For each test group, we report mean and standard deviation of these two metrics on the 25 runs.
We also report the overall averaged mean performance across the 10 test groups for each candidate feature representation.

\subsubsection{Results}
\begin{figure}[t]
\centering
\includegraphics[width=\linewidth]{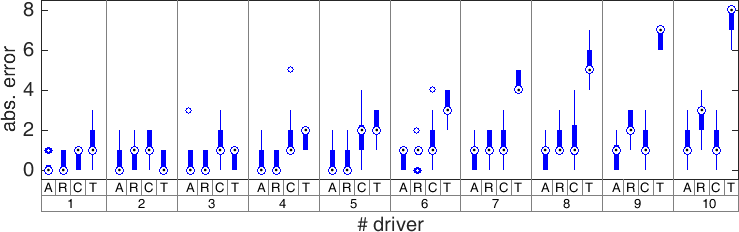}
\vspace{-0.28in}
\caption{Box plot of abs. error}
\label{fig:errboxplot}
\vspace{0.11in}
\includegraphics[width=\linewidth]{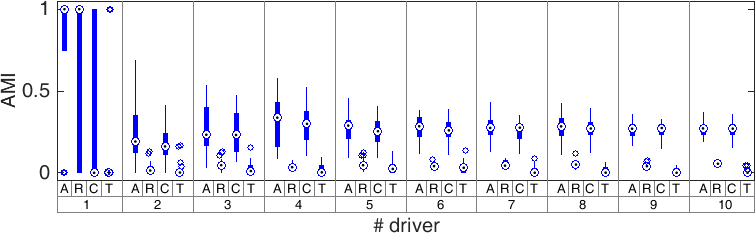}
\vspace{-0.28in}
\caption{Box plot of AMI score}
\label{fig:amiboxplot}
\vspace{-0.15in}
\end{figure}

Results are shown in Tables~\ref{tab:driverclusteringerror} and \ref{tab:driverclusteringami}.
The best entries in each test group (row) are bolded.
We can see that as the number of driver grows, the problem becomes harder.
Overall, clustering based on ARNet feature demonstrates the best performance among all.
Table~\ref{tab:driverclusteringerror} shows that ARNet feature leads to the mean error less than one driver in 9 out of the 10 tests.
It wins 5 out of the 10 tests with the least mean error, and places the second best in the 5 lost ones all by a small margin (0.12 at most).
Its averaged mean error of all the tests is just 0.68.
%The average of the mean AMI scores is 0.32, indicating an acceptable quality of clustering.
In contrast, the single-loss networks' features and the TripGBDT feature more often leads to larger errors. They all have the averaged mean error greater than 1.
Table~\ref{tab:driverclusteringami} shows that ARNet feature also leads to the best clustering quality.
It wins 9 (including a tie) of the 10 tests with the highest mean AMI, only with lost the 9-th test by 0.01.
Its averaged mean AMI 0.32 is the highest, while other candidates' are much worse.
We depict the box plots of the results in Figures~\ref{fig:errboxplot} and \ref{fig:amiboxplot}, where A, R, C and T stand for ARNet, RONet, CONet and TripGBDT features, respectively.
From the comparisons, we can conclude that the ARNet results are often significantly better.
The handcrafted TripGBDT feature performs the worst, implying the superiority of learning representations by deep networks. % over handcrafted ones.

\begin{table}[t]
\caption{Driver number estimation: abs. error}
\label{tab:driverclusteringerror}
\tiny
\centering
\begin{tabular}{c|c|c|c|c}
\hline
%\multirow{2}{*}{\# driver} &\multicolumn{4}{c|}{ARNet} & \multicolumn{4}{c|}{RONet} & \multicolumn{4}{c|}{CONet} & \multicolumn{4}{c}{TripGBDT} \\
%\cline{2-17}
\# driver & {ARNet} & {RONet} & {CONet} & {TripGBDT} \\
%& mean $\pm$ std &	mean $\pm$ std &	mean $\pm$ std &	mean $\pm$ std	\\
\hline
1 & \textbf{0.24} $\pm$ 0.43 	& 0.40 $\pm$ 0.49		& 0.56 $\pm$ 0.50		& 6.92 $\pm$ 27.4\\
2 & 0.48 $\pm$ 0.70 	& 0.76 $\pm$ 0.59		& 1.12 $\pm$ 0.71		& \textbf{0.44} $\pm$ 0.50\\
3 & 0.52 $\pm$ 0.70 	& \textbf{0.48} $\pm$ 0.50		& 1.24 $\pm$ 0.76		& 0.60 $\pm$ 0.49\\
4 & 0.52 $\pm$ 0.57 	& \textbf{0.40} $\pm$ 0.50		& 1.52 $\pm$ 0.90		& 1.52 $\pm$ 0.50\\
5 & 0.48 $\pm$ 0.57 	& \textbf{0.40} $\pm$ 0.57		& 1.72 $\pm$ 1.00 		& 2.32 $\pm$ 0.61\\
6 & \textbf{0.64} $\pm$ 0.48 	& 0.80 $\pm$ 0.50		& 1.40 $\pm$ 0.98		& 3.28 $\pm$ 0.53\\
7 & \textbf{0.80} $\pm$ 0.63 	& 1.32 $\pm$ 0.61		& 1.40 $\pm$ 0.89		& 4.48 $\pm$ 0.50\\
8 & \textbf{0.72} $\pm$ 0.72 	& 1.52 $\pm$ 0.57		& 1.52 $\pm$ 1.17		& 5.40 $\pm$ 0.69\\
9 & \textbf{0.92} $\pm$ 0.74 	& 2.40 $\pm$ 0.57		& 1.52 $\pm$ 0.70		& 50.7 $\pm$ 216$^*$\\
10& 1.44 $\pm$ 0.75		& 2.52 $\pm$ 0.75		& \textbf{1.36} $\pm$ 0.84		& 7.68 $\pm$ 0.55\\
\hline
avg & \textbf{0.68}
	& {1.10}
	& {1.34}
	& 8.34\\
\hline
\end{tabular}
%\vspace{-0.05in}
%\flushleft
\\
$^*$Huge outliers exist here, but the samples are not shown in Figure~\ref{fig:errboxplot} box plot's scope due to that displaying them in the graph will make the comparisons hard to read at the small scale.
\vspace{-0.07in}
%\end{table}
%
%
%\begin{table}[t]
\caption{Driver number estimation: AMI score}
\label{tab:driverclusteringami}
\tiny
\centering
\begin{tabular}{c|c|c|c|c}%{c|cccc|cccc|cccc|cccc}
\hline
%\multirow{2}{*}{\# driver} &\multicolumn{4}{c|}{ARNet} & \multicolumn{4}{c|}{RONet} & \multicolumn{4}{c|}{CONet} & \multicolumn{4}{c}{TripGBDT} \\
%\cline{2-17}
%& min & mean & max & std &			min & mean & max & std &		min & mean & max & std &		min & mean & max & std	\\
\# driver & {ARNet} & {RONet} & {CONet} & {TripGBDT} \\
\hline
1 &	\textbf{0.76} $\pm$ 0.47		& 0.60 $\pm$ 0.49		& 0.44 $\pm$ 0.50				& 0.12 $\pm$ 0.33\\
2 &	\textbf{0.24} $\pm$ 0.14		& 0.03 $\pm$ 0.03		& 0.19 $\pm$ 0.11				& 0.02 $\pm$ 0.04\\
3 &	\textbf{0.28} $\pm$ 0.14		& 0.05 $\pm$ 0.03		& 0.25 $\pm$ 0.12				& 0.03 $\pm$ 0.04\\
4 &	\textbf{0.31} $\pm$ 0.14		& 0.04 $\pm$ 0.02		& 0.29 $\pm$ 0.12				& 0.02 $\pm$ 0.03\\
5 &	\textbf{0.27} $\pm$ 0.09		& 0.05 $\pm$ 0.03		& 0.25 $\pm$ 0.09				& 0.03 $\pm$ 0.04\\
6 &	\textbf{0.28} $\pm$ 0.07		& 0.04 $\pm$ 0.02		& 0.25 $\pm$ 0.07				& 0.03 $\pm$ 0.04\\
7 &	\textbf{0.27} $\pm$ 0.07		& 0.05 $\pm$ 0.02		& 0.25 $\pm$ 0.06				& 0.02 $\pm$ 0.03\\
8 &	\textbf{0.28} $\pm$ 0.08		& 0.05 $\pm$ 0.02		& 0.27 $\pm$ 0.07				& 0.02 $\pm$ 0.02\\
9 &	0.26 $\pm$ 0.05		& 0.04 $\pm$ 0.01		& \textbf{0.27} $\pm$ 0.05					& 0.01 $\pm$ 0.02\\
10&	\textbf{0.27} $\pm$ 0.04		& 0.05 $\pm$ 0.02		& \textbf{0.27} $\pm$ 0.04		& 0.01 $\pm$ 0.01\\
\hline
avg & \textbf{0.32}
	& {0.10}
	& {0.27}
	& 0.03\\
\hline
\end{tabular}
\vspace{-0.03in}
\end{table}

\begin{figure}[t]
\centering
\includegraphics[width=0.77\linewidth]{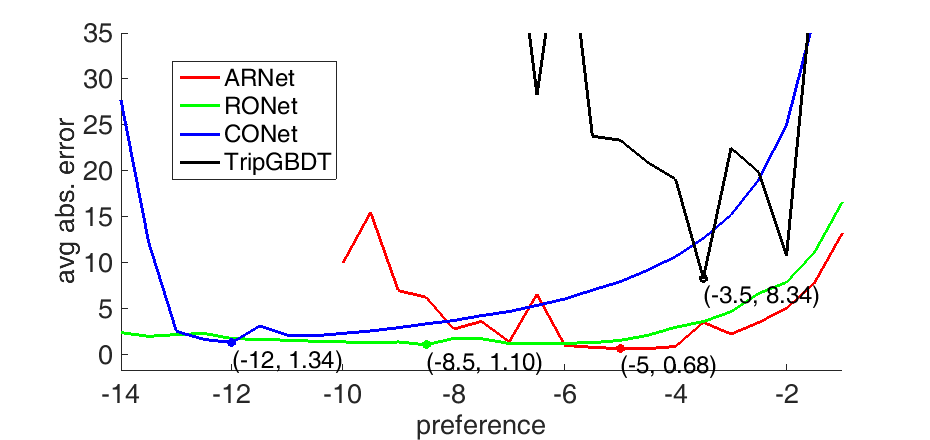}
\vspace{-0.15in}
\caption{Averaged mean abs. error vs. AP preference. The least-error points are highlighted for each curve.}
\label{fig:tunepreference}
\vspace{-0.15in}
\end{figure}

We studied how the estimation error changes with the AP preference setting, as shown in Figure~\ref{fig:tunepreference}.
We can see that the chosen thresholds lead to the best performance for each candidate, revealing that the advantage of ARNet feature is not due to the thresholding of clustering.

\begin{figure}[tb]
\centering
\subfigure[ARNet on seen data]{
\label{fig:multitask_seen_tSNE}
\includegraphics[width=0.47\linewidth]{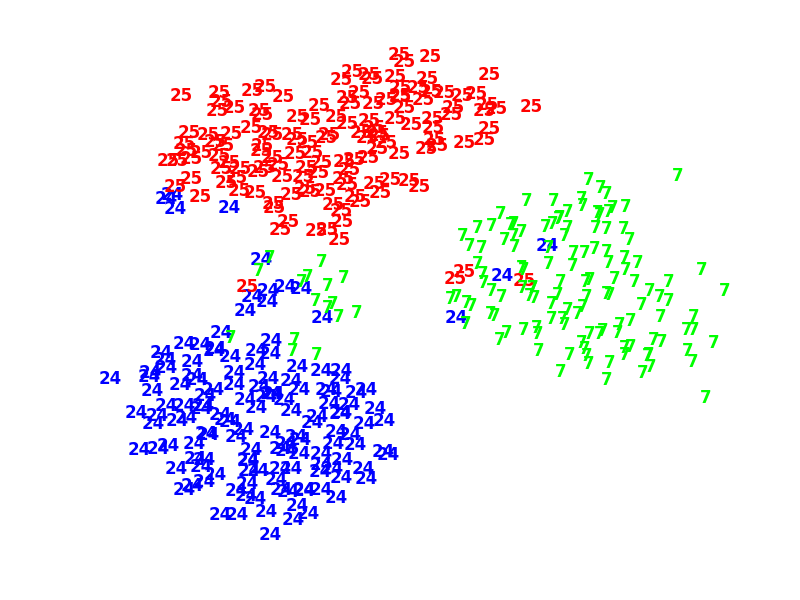}
}
\subfigure[ARNet on unseen data]{
\label{fig:multitask_unseen_tSNE}
\includegraphics[width=0.47\linewidth]{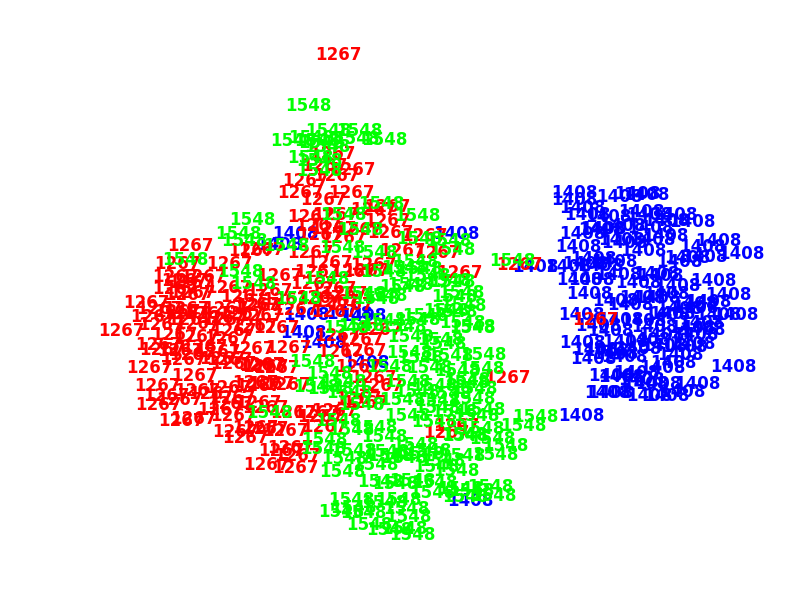}%{multiTask150_2014_2505.png}
}
\\
\vspace{-0.15in}

\subfigure[RONet on seen data]{
\label{fig:reconstruction_seen_tSNE}
\includegraphics[width=0.47\linewidth]{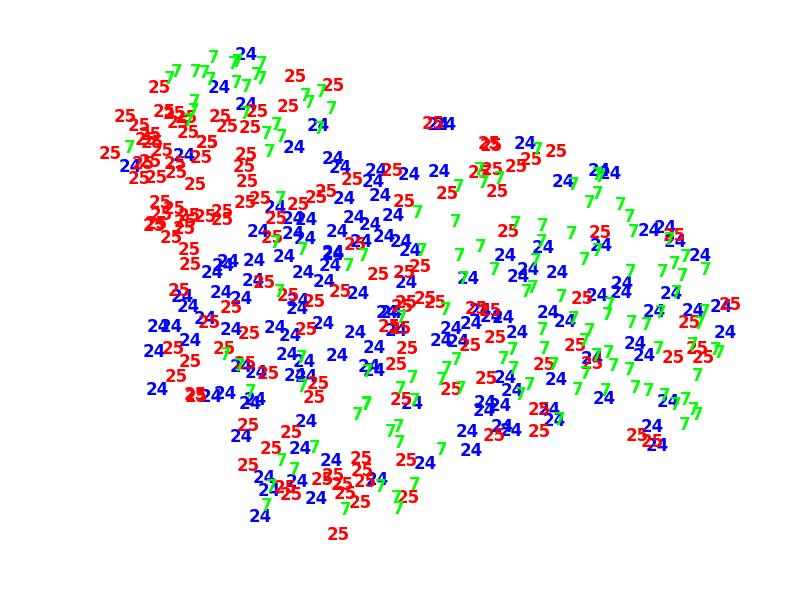}
}
\subfigure[RONet on unseen data]{
\label{fig:reconstruction_unseen_tSNE}
\includegraphics[width=0.47\linewidth]{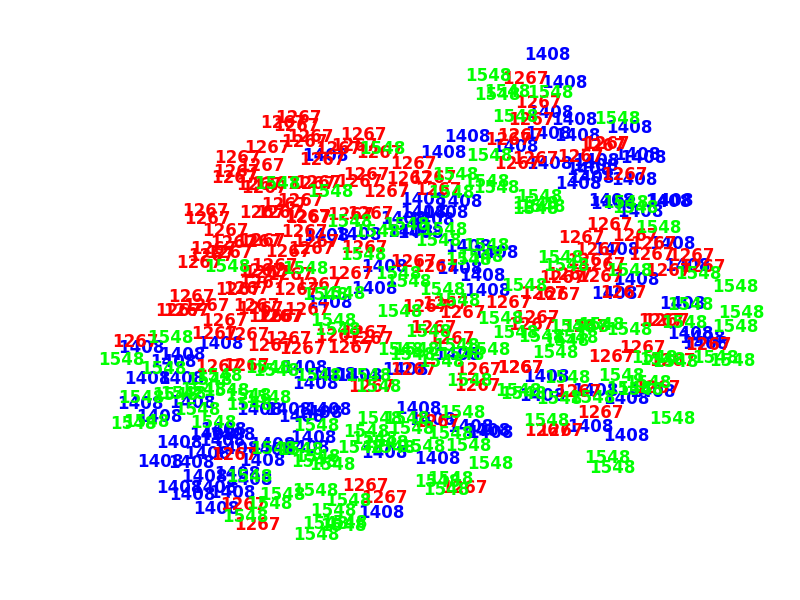}%{rec150_2014_2505.png}
}
\\
\vspace{-0.15in}

\subfigure[CONet on seen data]{
\label{fig:classification_seen_tSNE}
\includegraphics[width=0.47\linewidth]{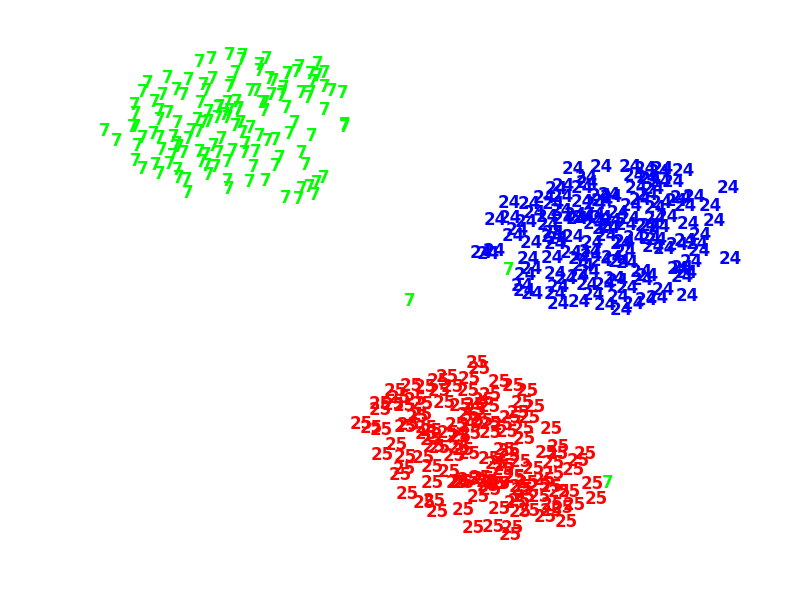}
}
\subfigure[CONet on unseen data]{
\label{fig:classification_unseen_tSNE}
\includegraphics[width=0.47\linewidth]{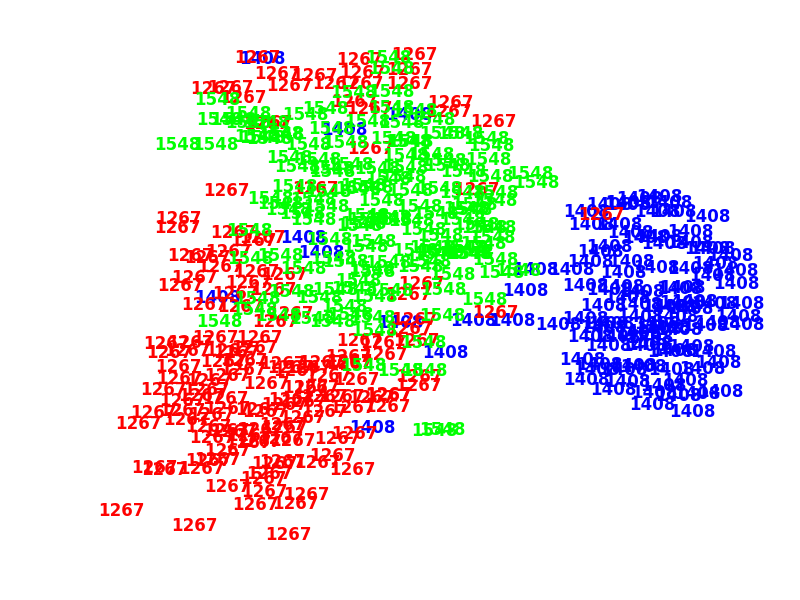}%{clf_150_2014_2505.png}
}
\\
\vspace{-0.15in}

\subfigure[TripGBDT on seen data]{
\label{fig:tripgbdt_seen_tSNE}
\includegraphics[width=0.47\linewidth]{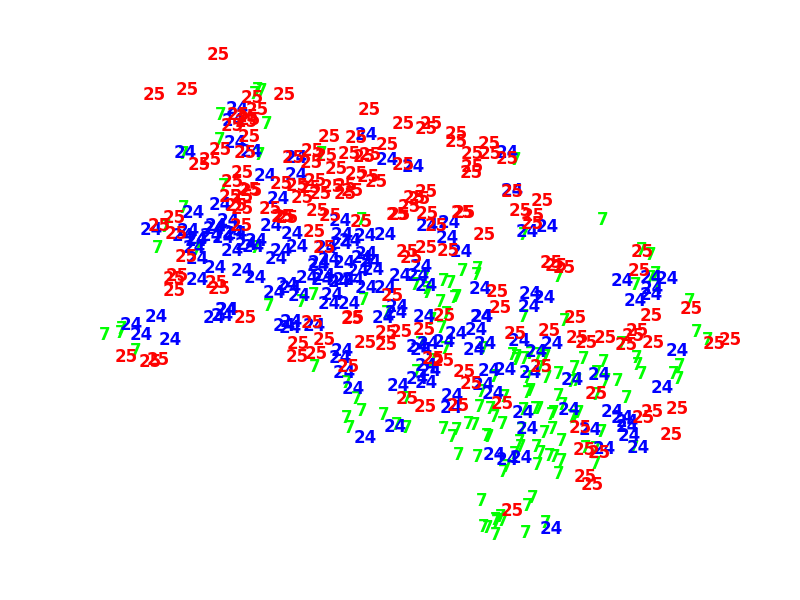}
}
\subfigure[TripGBDT on unseen data]{
\label{fig:tripgbdt_unseen_tSNE}
\includegraphics[width=0.47\linewidth]{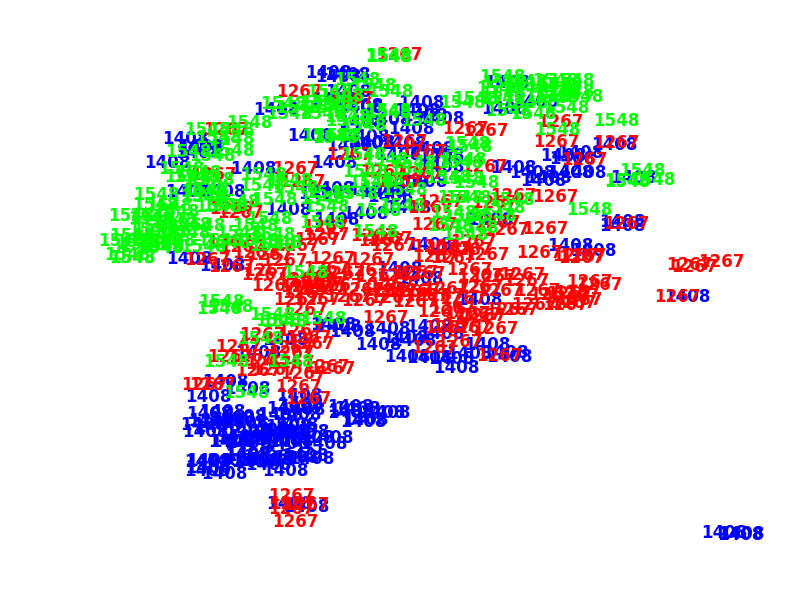}
}
\vspace{-0.17in}

\caption{t-SNE projections of trip features, labeled by driver ID.
Left column: results on training trips from 3 seen drivers (ID: 7, 24, and 25).
Right column: results on trips from 3 unseen drivers (ID: 1267, 1408, and 1548).% in one of the 3-driver tests.
%The second row are results on three unseen drivers (ID: 161, 1222, and 1260) from one of the testing groups.
}
\label{fig:tSNE}
\vspace{-0.12in}
\end{figure}

We further use t-SNE \cite{maaten2008visualizing} to project the trip feature representations onto a 2-d space for visual comparisons.
Again, we employ the scikit-learn \cite{scikit-learn} t-SNE implementation with all parameters kept default.
Figure~\ref{fig:tSNE} shows typical results on both seen and unseen drivers.
We can see that the ARNet feature is robust: Trips of a same driver show relatively similar driving styles to each other, exhibiting clear clustering patterns, no matter the drivers are seen or unseen.
In contrast, due to the absence of supervisory information, RONet learns ``too generalized'' feature that cannot differentiate drivers well, leading to poor clustering quality (small AMI scores in Table~\ref{tab:driverclusteringami}).
The CONet feature is highly discriminative on seen trips and drivers. Nonetheless, on unseen drivers, it more easily split same-class trips faraway (driver ID 1267 in red color).
Similar to RONet result, the TripGBDT feature does not reflect clear clustering patterns, explaining its poor performance.
In a word, ARNet learns a better generalized driving style representation, % than the single-task ones,
resulting in interpretable better performance on the driver number estimation problem.

\subsection{Driver Identification Problem}
We now study the classical driver identification problem to demonstrate that ARNet also helps to improve the driver identification (i.e., classification) accuracy.
Based on the same 50 drivers used in training on the driver number estimation problem, we train different candidate models and compare both segment-level and trip-level driver classification accuracies (50-class), where the predictions indicate to which driver the segment/trip belongs.
A trip-level prediction is obtained by summing up the predictions on segments of this trip to give a vote weighted by the confidence scores.
For each driver, we use 80\% trips for training and the rest 20\% for testing.
Note that here the testing trips are all from seen drivers, though the trips are excluded in training.

\subsubsection{Candidate Methods}
In addition to ARNet and CONet, % in Figure~\ref{fig:classification_only_architecture},
we compare five different supervised learning deep networks, NoPoolCNN, CNN, PretrainIRNN, IRNN and StackedIRNN studied in \cite{dong2016characterizing} and two GBDT methods, the TripGBDT and a GBDT reading the $35\times 128$ input as a flattened feature vector for learning. %:
%\vspace{-0.05in}
%\begin{itemize}
%  \item CNN: a 6-layer convolutional neural network.% using 1-d convolution and pooling.
%  \vspace{-0.05in}
%  \item NoPoolCNN: CNN without pooling layers.%, to show that pooling is as important as convolution.
%  \vspace{-0.05in}
%  \item IRNN \cite{le2015simple}: an RNN using the identity matrix for recurrent weight matrix initialization.%, with 100 hidden units in the recurrent layer.
%  \vspace{-0.05in}
%  \item PretrainIRNN: an IRNN trained by using the features extracted from the pre-trained CNN as inputs.
%  \vspace{-0.05in}
%  \item StackedIRNN: a stacked IRNN with 2 recurrent layers.%, with 100 hidden units in each recurrent layer.%, to figure out how the network's complexity affect the performance.
%  \vspace{-0.05in}
%  \item GBDT \cite{friedman2001greedy}: directly reads the $35\times 128$ input as a flattened 4480-d feature vector as input.
%  \vspace{-0.05in}
%  \item TripGBDT \cite{dong2016characterizing}: GBDT running on the 57-d handcrafted feature representation.
%  \vspace{-0.05in}
%\end{itemize}
%Results of these methods are adopted from \cite{dong2016characterizing}, where readers can find more details. % of the algorithms.
%The same protocols are applied to training
ARNet and CONet settings are kept unchanged as in the driver number estimation experiments.
In ARNet, the fc3 output is used for prediction.
%We directly compare the classification loss
%uses the same parameter settings as in the driver number estimation problem.

\subsubsection{Results}

\begin{table}[tb]
\caption{Driver identification accuracy (in \%)}\label{tab:driveridentification}
\centering  % 表居中
\tiny
\begin{tabular}{l|c|c|c}  % {lccc} 表示各列元素对齐方式，left-l,right-r,center-c
\hline
%\vspace{1mm}
method &segment &trip top-1 &trip top-5\\
\hline%\hline  % \hline 在此行下面画一横线
NoPoolCNN &16.9 &28.3 &56.7\\
CNN &21.6 &34.9 &63.7\\
PretrainIRNN &28.2 &44.6 &70.4 \\
IRNN &34.7 &49.7 &76.9\\
StackedIRNN &{34.8} &{52.3} &{77.4}\\
%\hline
GBDT &18.3 &29.1 &55.9\\
TripGBDT &- &51.2 &74.3\\
%\hline
CONet & 37.5 & 56.1 & 74.9\\
ARNet & \textbf{40.4} & \textbf{58.2} & \textbf{78.3}\\
\hline
%\vspace{1mm}
\end{tabular}
\vspace{-0.1in}
\end{table}

Table~\ref{tab:driveridentification} summarizes both segment and trip level accuracies.
ARNet outperforms all the other candidates with the highest accuracies, segment 40.4\%, trip top-1 58.2\%, and trip top-5 78.3\%.
It improves the accuracies by roughly 3\% compared with CONet, which performs the second best in terms of segment and trip top-1 performance.
This verifies that ARNet also helps learn a better classification feature representation that improves the supervised learning performance.

\section{Related Work}
Existing approaches on driving style learning usually follow the supervised learning paradigm, whether or not the input is GPS data.
Many methods based on non-deep-learning classifiers and reinforcement learning were proposed, e.g., by \cite{lopez2012driver,quintero2012,quek2013driver,van2013driver,kuderer2015learning}.
Recently, \cite{dong2016characterizing} extended deep learning to GPS data and proposed several CNNs and RNNs that can learn interpretable driving style features. But still, these are typical supervised classification networks.

Though in literatures, there are plenty of neural networks combining supervised and unsupervised learning, e.g., \cite{lee1992,karayiannis1997,raina2007self,collobert2008unified}, few attempts were made on employing the autoencoder structure as a special regularizer to supervised learning as in ARNet.
Especially, the autoencoder in ARNet is not for pre-training or applying unsupervised/supervised learning in turn, but for guiding the feature learning simultaneously with supervisory signals.

ARNet can also be viewed as a special case of multi-task learning (MTL) if regarding classification and reconstruction as two tasks.
However, MTL typically aims to learn a shared representation across tasks \cite{Caruana:1997:ML:262868.262872,Argyriou2007multi}, and in most (if not all) cases, the tasks are either all supervised or all unsupervised.
The only MTL method combining unsupervised reconstruction with supervised learning that we are aware of is the Semi-supervised Autoencoder for Multi-task Learning (SAML) \cite{zhuang2015representation}.
%However, SAML and ARNet are essentially different. % due to different goals.
In SAML, %the goal is the traditional classification accuracy.
each task is a combination of reconstruction and classification,
and a shared autoencoder reconstructs the network input via shared feature layers used also for classification.
Most differently,
%The most significant difference between SAML and ARNet is that,
%the goal is to learn generalized feature representations for unseen classes not observed in training.
%classification and reconstruction can be regarded as two tasks, not a combined one, and
the autoencoder in ARNet reconstructs the shared hidden-layer feature instead of the network input.
Also, the two tasks in ARNet (reconstruction and classification) use different (fc1 and fc3, respectively) but not a shared representation.
Besides, SAML is not designed for learning from GPS data.
To the best of our knowledge, ARNet is the first attempt on combining unsupervised autoencoder and supervised learning in a unified deep architecture for learning from GPS data.

%\section{Discussion}
%
%if transfer learning can be applied

\section{Conclusion}
In this paper, we study learning driving style representation from GPS data, and propose a novel deep architecture, Autoencoder Regularized Network (ARNet), which combines unsupervised and supervised feature learning by introducing an autoencoder as a special regularizer to supervised feature learning.
ARNet can also be viewed as adding supervisory information to unsupervised feature learning of the autoencoder.
In both ways, it improves the quality of learnt driving style representation.
We further propose trip2vec, a trip encoding framework using ARNet as the base encoder to extract trip-level driving styles.
Experiments on benchmark problems verify the advantages of ARNet over existing methods, especially on characterizing new drivers.
%Besides the two studied driver number estimation and driver identification problems, ARNet can also help solving problems such as to early detect abnormal driving status (e.g., drunk, fatigue, and drowsy), which will be our future work.
Future work includes studying the performance of ARNet on other related problems such as to early detect abnormal driving status (e.g., drunk, fatigue, and drowsy) and those representation learning problems in other domains.

\clearpage
\small

\bibliographystyle{named}
\bibliography{DeepLearning}

\end{document}